\documentclass[10pt, a4paper]{article}
\usepackage{lrec}
\usepackage{amsmath}
\usepackage{amssymb}
\usepackage{stfloats}
\usepackage[linesnumbered]{algorithm2e}
\usepackage{graphicx}
\usepackage{multirow}
\usepackage{times}
\usepackage{url}
\usepackage{makecell}
\usepackage{latexsym}
\usepackage[utf8]{inputenc}
\usepackage{booktabs}
\usepackage[labelfont=bf]{caption}
\newcommand*\mean[1]{\bar{#1}}

\DeclareMathOperator*{\kargmax}{k~arg\,max}

\DeclareMathOperator*{\kargmin}{k~arg\,min}

\title{Distributional Term Set Expansion}
\name{Amaru Cuba Gyllensten, Magnus Sahlgren}
\address{RISE SICS\\Box 1263, SE-164 29 Kista \\ 
\{amaru.cuba.gyllensten, magnus.sahlgren\}@ri.se}
\date{}

\abstract{
This paper is a short empirical study of the performance of centrality and classification based iterative term set expansion methods for distributional semantic models. Iterative term set expansion is an interactive process using distributional semantics models where a user labels terms as belonging to some sought after term set, and a system uses this labeling to supply the user with new, candidate, terms to label, trying to maximize the number of positive examples found. While centrality based methods have a long history in term set expansion \cite{sarmento2007more,pantel2009web}, we compare them to classification methods based on the the Simple Margin method, an Active Learning approach to classification using Support Vector Machines \cite{tong2002support}. Examining the performance of various centrality and classification based methods for a variety of distributional models over five different term sets, we can show that active learning based methods consistently outperform centrality based methods.\\ 
\newline 
}

\begin{document}

\maketitleabstract

\section{Introduction}

One of the 
most commonly used resources in Natural Language Processing is the \emph{term set}: a set of, optionally, labeled words. It is a standard approach to sentiment, subjectivity, and stance detection: compile lists of terms representing the categories in question, and then calculate the occurrence of these terms in data. A text is assigned to the category whose terms are most prevalent in the text. This approach -- often referred to as {\em lexicon-based classification} -- is simplistic, but surprisingly powerful, and often provides useful results in the absence of supervised classifiers \cite{Eisenstein17}. Another closely related use case is topic monitoring in social media, in which case the frequency of topic-related terms over time can be used to gauge public interest in those topics. 


The performance of such lexicon-based approaches obviously depends on the quality of the lexicon being used. A common approach is to use distributional models (word embeddings) to populate the lexicon on the basis of a small set of manually selected seed terms (e.g.~``bad" and ``subpar" as seed terms for negative sentiment, and ``good" and ``ace" as seed terms for positive sentiment). The seed terms are used as probes into the distributional model with the goal of finding other terms that are (distributionally) similar to the seed terms. 
Iterative term set expansion is the iterated, interactive, version of this procedure: An annotator defines an initial, incomplete, term set. This term set is fed to the term set expansion method, generating new candidate terms. The annotator labels these as belonging to, or not belonging to, the term set, which is updated accordingly. The new updated term set is then fed to the expansion method, and the process is repeated indefinitely.
In this way, a small set of manually defined seed terms can be (semi-) automatically expanded into a potentially very large lexicon.

Expanding term sets using distributional models usually amounts to computing the similarity between all terms in the model and the seed terms, and then including the candidates that are most similar to the seed terms. This may seem like a well-defined process, but the quantification of similarity can be done in many different ways, and the choice of similarity function will have a significant impact on the quality of the resulting lexicon. To the best of our knowledge, there are no published comparisons between different ways of expanding term sets using distributional models, and consequently, we still lack a best practice for distributional term set expansion.

This paper aims to fill this void. In the following sections, we compare a number of standard approaches for iterative distributional term set expansion, with the aim of identifying a best practice for using distributional models to expand term sets. In doing so, we provide answers to the following questions: which methods are commonly used for term set expansion using distributional models? What are the performance differences between these methods? Is any of the methods more suitable to use for specific distributional models? And finally, is any method superior in general (and could consequently be described as a best practice)?

\section{Distributional models}

The quality of a distributionally-derived lexicon for classification purposes also depends on the choice of distributional model. We include the standard types of distributional models, which are detailed in the following sections, in our experiments.

\subsection{(Weighted) Count models}
The simplest distributional models are count-based models: for some notion of target and context items one counts the number of times the context item co-occurred with each target item. These models can be extended with weighting schemes to better fit the problem at hand. The most widely used and studied are variants on Pointwise Mutual Information (PMI), such as Positive PMI (PPMI), Smoothed PPMI, and Shifted PPMI \cite{levy2015improving}.

\subsection{Factorized count models}
A common method to speed up usage and computation on distributional models is to factorize a (weighted) count matrix using truncated Singular Value Decomposition (SVD). Recently it has been shown that one can greatly improve model performance by altering the singular values of the singular value decomposition, such as taking the square root of each singular value, or dropping them completely. In this study we have 
opted for the square root of the singular values, based on the results in \cite{levy2015improving}.

\subsection{Prediction models}
The two prediction based models used are SkipGram with Negative Sampling (SGNS), and Continuous Bag Of Words (CBOW) \cite{Mikolov:2013}. SGNS strives to predict whether an observation (consisting of a target word and a context word among the surrounding words) came from the data or was sampled from a distribution of negative examples. The objective of CBOW is the same, but instead of predicting each target-context pair, CBOW averages over all context items for the given observation.

\subsection{Model choice}
Ultimately, the models chosen were Factorized PPMI, Factorized Smoothed PPMI, SGNS, and CBOW, all with a window size of 2, and dimension 200. 
All models were trained on text data from the British National Corpus \cite{Clear:1993:BNC:166403.166418}.

\section{Iterative Term Set Expansion Methods}

Iterative Term Set Expansion is the method of iteratively, with user input, expanding a term set. In this paper we have formalized it in the following way:  

Given a labeling function $label : \text{Term} \rightarrow \text{Label}$\footnote{$f : a$ denotes $f$ has type $a$, $a \rightarrow b$ is the type of functions from $a$ to $b$, $a\times b$ denotes the types of pairs of variables $a$ and $b$, and $[a]$ denotes a finite set of $a$s, Label is, in this work, always taken to be boolean (i.e. True if the term is in the term set and False otherwise), and Term is, rather sloppily, used to refer both to the actual term and its distributional representation.}
(which would be a human annotator), an expansion method $expand : [\text{Term} \times \text{Label}] \rightarrow [\text{Term}]$, and a set of already labeled terms $L_t : [\text{Term} \times \text{Label}]$, the labeled terms $L_t$ are fed to the expansion method $expand$, which gives a set of new candidate terms to be labeled by the labeling function (or human annotator) $label$, resulting in a larger, and hopefully more informative, set of labeled terms $L_{t+1}$. If $L_0$ is the initial term set, $L_i$ is the result after $i$ expansion-labeling steps.\footnote{This can be expanded to the case where $expand$ returns a stream of terms to be labeled, and $label$ can decide to label, skip, or demand a new expansion with the recently labeled data, but that has been left out for simplicity's sake.}
\begin{equation}
L_{t+1} = L_t \cup \{(x, label(x)) | x\in expand(L_t)\}
\label{eq:expand}
\end{equation}

Methods used to find candidate terms (the $expand$ method in Equation (\ref{eq:expand})) can be characterized as either \emph{centrality} based or \emph{classification} based. Centrality based methods work by constructing a representation of the term set within the distributional model. In essence constructing a synthetic, \emph{central}, proxy term, whose neighborhood is taken to be representative of the whole term set. 
Centrality based methods have the advantage that one iteration of the term set expansion has complexity proportional to computing the central representation and performing a neighborhood query in the distributional model. Both of which are usually very quick operations with even more potential speed up if the centrality computation can be done in a streaming and/or parallel fashion.
Classification based methods work instead by constructing a classifier based on the term set, i.e.,~a function from the distributional model's underlying space to some measure of belonging to the given term set. As such they are a superset of centrality based measures, where the measure of belonging to the term set is the similarity to its central representation.

\subsection{Centrality based methods}
The most intuitive centrality based method is the {\bf centroid expansion method}: given a term set, its central representation is the average of all term vectors in the set:
\begin{equation}
\text{centroid}(T) = \mean{T} = \frac{1}{|T|}\sum_{t\in T} t.
\end{equation}
Apart from being an intuitive and familiar notion of centrality, it also has the property that similarity to the centroid $\mean{T}$ is equivalent to the average similarity to terms in $T$, if similarity is an inner product on the vector space.

As a slight modification to the centroid method we introduce a simple {\bf Signal-to-Noise ratio} centroid: the central representation of a term set is the average of all term vectors divided by empirical standard deviation:
\begin{equation}
\text{snr}(T) = \mean{T} \sqrt{\frac{1}{|T|-1}\sum_{t\in T}(t - \mean{T})^2}.
\end{equation}
The intuition is to scale down the importance of noisy dimensions and scale up the importance of dimensions where there is less noise.

{\bf Eigencentrality} is a centrality measure usually associated with graphs, most famously used by Pagerank to rank importance of webpages \cite{Pageetal98}. Given an adjaceny matrix $A$, the eigencentrality is given by the eigenvector of $A$ with the largest eigenvalue. 
Here, we compute the eigencentrality $W^TW$, where $W$ is the $\text{Term}\times\text{Feature}$ matrix of the term set, and use the resulting vector as the central representation of the term set. This scales up the importance of central terms in the term set, and scales down the importance of peripheral terms.

To find new candidate terms to be labeled, we have chosen to return the unlabeled terms closest to the central representation of the positive examples in the term set:

\begin{align*}
expand_{center}(L) &= \kargmax_{t\in \text{Vocab} \setminus L} sim(t, c) \\
\text{where}~ L^+ &= \text{positive examples in } L \\ 
c &= center(L^+)
\end{align*}

\subsection{Classification based methods}
Classification methods are more general than centrality based methods, and where centrality based measures lack simple means of accounting for negative examples, negative examples are crucial for a classification based approach. In a classification setting, the challenge is not about leveraging information from negative examples, but how to deal with the sparsity of labeled examples. One solution to this problem is Active Learning \cite{olsson2009literature}.

Active learning is a subfield of machine learning that incorporates the selection and labeling of data into the learning framework. In this case, the active learner is used to train a classifier based on the labeled points, \emph{and} to suggest new data points to label such that these new data points are as informative as possible for the active learner.

In this paper we have restricted ourselves to Support Vector Machines (SVMs), since these admit simple and efficient methods for active learning. We use {\bf RBF kernels} and {\bf Linear kernels} for the SVM. The motivation behind using Linear kernels is their simplicity, and the fact that all centrality based methods can be subsumed by linear classifiers\footnote{
Using the definition of centrality based methods we've used here, and assuming that notion of distance in the distributional space is an inner product, then the resulting measure of belonging of all centrality based methods are interchangeable with a linear classifier.
}
. The motivation behind using RBF kernels is, apart from their ubiquity, the fact that they capture the local influence of the supplied examples.

For both classification methods we have used the Simple Margin method to find new candidate terms. Simple Margin uses the structure of Support Vector Machines to select informative data points. This works by choosing the $k$ words closest to the separating hyperplane as the next ones to be labeled:

\begin{align*}
expand_{margin}(L) &= \kargmin_{t\in \text{Vocab} \setminus L} |d(t)| \\
\text{where}~d(t) &= classify(L)(t)
\end{align*}

The intuition here being that these are the data points the algorithm is the most unsure of, and whose \emph{minimum} influence on the loss function is maximal \cite{tong2002support}. Note that this is not designed to maximize the number of positive examples we supply the labeler, but for the labeling of candidate terms to be as informative as possible for the underlying classifier. 

\section{Experimental setup}

What we want to find out is how informative a tool such as this could be to a human annotator when building a term set. As such we are interested in how many positive examples the expansion method supplies the labeler with per iteration. This was evaluated against a number of predefined term sets: positive and negative sentiment term sets extracted from the AFINN word list \cite{nielsen2011new} (an \emph{affective word list}), the elements term set from Pantel \cite{pantel2009web}, a color term set extracted from Wikipedia, and an ingredient term set extracted from Wikibooks cookbook.

These predefined term sets are used as proxies for a human annotator: Given a term set, we construct a random initial labeled sets with five terms taken from the term set, and five terms taken at random from terms not in the term set. When the iterative term set expansion queries the annotator for a label, this label is extracted directly from the predefined term set, i.e. with a labeling function defined as in Equation \ref{eq:label}.

\begin{equation}
\label{eq:label}
label_D(x) =
\begin{cases}
	Positive&,x \in D \\
    Negative&,x \not \in D 
\end{cases}
\end{equation}
\begin{figure}
\footnotesize  
\centering
\begin{tabular}{|l|c|c|}
\multicolumn{1}{c|}{} & \thead{\textbf{Positive examples}} & \thead{\textbf{Negative examples}} \\
\hline
$L_0$ & 
\makecell{responsive, perfects, \\ popular, opportunity, \\ comforting} & 
\makecell{acropolis, bogus, contestants, \\ tartuffe, counter-themes} \\
\hline
\multicolumn{3}{c}{\makecell{
 $expand(L_0)$: \\
 \underline{agreeable}, \underline{supportive}, adaptable, attentive, conducive, \\
 non-threatening, open-minded, receptive, self-critical, sociable
}} \\
\hline 
$L_1$ & 
\makecell{responsive, perfects, \\ popular, opportunity, \\ comforting, \underline{agreeable} \\ \underline{supportive}} & 
\makecell{
acropolis, bogus, contestants, \\
tartuffe, counter-themes, \\
attentive, adaptable, conducive, \\
non-threatening, open-minded, \\
receptive, self-critical \\
sociable
} \\
\hline 
\multicolumn{3}{c}{\makecell{
 $expand(L_1)$: \\
\underline{encouragement}, 
\underline{reassuring},
\underline{support}, instant, invaluable \\
np, reassurance, salutary, snp, thatcher
}} \\
\multicolumn{3}{c}{\textbf{\vdots}}
\end{tabular}
\caption{
Example of the iterative term set expansion process.
Starting out with an initial labeled term set $L_0$ consisting of five positive and five negative examples of the sought after term set, we expand $L_0$ to get ten candidate terms ($expand(L_0)$). Of these ten candidate terms, the annotator labels ``agreeable" and ``supportive" as belonging to the sought after term set, and the labeled term set is updated accordingly. The procedure is then repeated with the updated term set $L_1$, yielding ten new candidates which the annotator labels, and so on, until a satisfactory term set has been constructed.
In this particular case, the sought after term set is the AFINN POS term set, with the ``annotator" being a simple lookup as described in Equation \ref{eq:label}.
The performance of the term set expansion method would be the average number of positive examples added to the labeled term set, in this case $2.5$.
}
\label{fig:example}
\end{figure}

Each combination of distributional model and expansion method was evaluated by running the term set expansion procedure for twenty steps, querying the ``annotator" to label ten candidate terms at each step, for ten random initial labeled term sets\footnote{The initial term sets were shared across models and methods.}. The reported performance is the average number of positive examples among the candidate terms per iteration. An example of the first step of this procedure can be seen in Figure \ref{fig:example}: The initial labeled term set $L_0$, sample from AFINN POS, is expanded, labeled, updated, and expanded again.

\section{Results}

\begin{table}
\footnotesize  
\centering
\begin{tabular}{lrrrr}
\toprule
{} & \multicolumn{4}{l}{\large{Ingredients}} \\
 &        CBOW &  PPMI &  \textbf{SGNS} & SPMI \\
\midrule
centroid expansion   &        1.69 &              2.35 &  1.17 &            2.37 \\
eigencentrality      &        1.25 &                  1.64 &  0.68 &            1.32 \\
signal to noise      &        1.45 &                  1.72 &  0.67 &            0.82 \\
simple margin linear &        0.36 &                   2.31 &  1.68 &            1.58 \\
\textbf{simple margin rbf}    &        1.15 &                  2.46 &  \textbf{2.47 }&            1.49 \\
\end{tabular}

\begin{tabular}{lrrrr}
\toprule
{} & \multicolumn{4}{l}{\large{Colors}} \\
 &   CBOW & PPMI &  \textbf{SGNS} & SPMI \\
\midrule
centroid expansion   &   1.53 &               1.92 &  0.38 &            1.47 \\
eigencentrality      &   0.62 &                   0.94 &  0.18 &            0.95 \\
signal to noise      &   0.81 &                  1.22 &  0.12 &            1.47 \\
simple margin linear &   0.67 &                    2.84 &  2.57 &            2.27 \\
\textbf{simple margin rbf}    &   3.20 &                    2.95 &  \textbf{3.47} &            2.58 \\
\end{tabular}

\begin{tabular}{lrrrr}
\toprule
{} & \multicolumn{4}{l}{\large{AFINN POS}} \\
 &      CBOW & \textbf{PPMI} &  SGNS & SPMI \\
\midrule
centroid expansion   &      1.19 &                     1.91 &  0.09 &            1.62 \\
eigencentrality      &      0.53 &                    0.98 &  0.05 &            0.96 \\
signal to noise      &      0.96 &                     1.78 &  0.08 &            1.53 \\
simple margin linear &      0.90 &                      3.63 &  2.51 &            2.33 \\
\textbf{simple margin rbf}    &      3.25 &                    \textbf{4.27} &  3.99 &            2.72 \\
\end{tabular}

\begin{tabular}{lrrrr}
\toprule
{} & \multicolumn{4}{l}{\large{AFINN NEG}} \\
 &      CBOW & PPMI &  \textbf{SGNS} & SPMI \\
\midrule
centroid expansion   &      2.97 &                    4.06 &  0.74 &            3.39 \\
eigencentrality      &      1.27 &                  2.09 &  0.33 &            1.94 \\
signal to noise      &      2.35 &                  3.32 &  0.21 &            3.03 \\
simple margin linear &      1.38 &                    3.79 &  3.68 &            3.20 \\
\textbf{simple margin rbf}    &      4.63 &                  4.59 &  \textbf{4.78 }&            3.52 \\
\end{tabular}

\begin{tabular}{lrrrr}
\toprule
{} & \multicolumn{4}{l}{\large{Elements}} \\
 &     CBOW &  PPMI &  SGNS & \textbf{SPMI} \\
\midrule
centroid expansion   &     1.77 &                     2.16 &  1.12 &            2.21 \\
eigencentrality      &     1.22 &                    1.74 &  0.74 &            1.67 \\
signal to noise      &     0.79 &                    2.09 &  0.52 &            2.08 \\
simple margin linear &     0.68 &                    2.59 &  1.52 &            2.58 \\
\textbf{simple margin rbf}    &     1.31 &           2.60 &  1.83 &   \textbf{2.65} \\
\bottomrule
\end{tabular}

\caption{Average number of positive examples found per iteration of the term set expansion method, based on ten random initialization with five positive and five negative examples. Simple Margin using an RBF kernel is consistently the best expansion method for all term sets.}
\label{tab:expansion}
\end{table}

Table \ref{tab:expansion} shows the average performance as described in the previous section, i.e. the average number of positive examples found per iteration. The results are displayed per tested term set, expansion method, and underlying distributional model used.
It is evident from this table that, generally, Simple Margin using an RBF-kernel outperforms the other expansions methods. This is true for all term sets, and almost all combinations of term sets and distributional models tested.

It is also evident that centroid expansion clearly outperforms the other centrality based expansion methods, and in some instances, for some models, outperforms Simple Margin with a linear kernel. 

\section{Conclusion \& discussion}

As a best practice when using distributional methods for term set expansion, our results indicate that simple margin using an RBF-kernel is the best choice for all term sets, regardless of the distributional model used. Simple Margin with a linear kernel -- which has both the advantage of being directly representable in the vector space, and being efficient to compute -- also consistently performed well for all distributional models but CBOW.

\begin{figure}
\footnotesize  
\centering
\begin{tabular}{c}
\toprule
\makecell{
\underline{strong},
enjoyed,
\underline{excited},
\underline{excellent},
tremendously, \\
\underline{thanks},
marvellous,
\underline{rich},
\textbf{disappointed},
uplifting,
\underline{fun},
\underline{enjoy}, \\
\underline{interesting},
\underline{enjoying},
\underline{healthy},
\underline{terrific},
\underline{lovely},
\underline{ambitious},\\
\underline{fantastic},
enjoyable,
\textbf{worried},
\underline{interested},
\textbf{sorry},
\underline{improved},\\
wonderfully,
\underline{powerful},
\textbf{upset},
\underline{successful},
\underline{relieved},
\underline{amazing},
}\\
\bottomrule
\end{tabular}
\caption{Top 30 unlabeled candidates for AFINN POS using PPMI and Simple Margin with an RBF kernel after an expansion procedure as described in the section 4. The underlined words are those in the predefined term set, and the bold words are words that we deemed erroneous}
\label{fig:classifier}
\end{figure}

It should be noted that, apart from providing the labeler with candidate terms, the simple margin methods also provides a classifier based on the labeled set. This could be used to quickly expand the term sets -- without supervision, but with some uncertainty -- to include all terms the classifier would consider positive examples. An example of this can be seen in Figure \ref{fig:classifier}.

Both SGNS and CBOW stands out: SGNS, while outperforming most other models when using the RBF method, performed terribly in conjunction with centrality based methods. For CBOW, there is a significant loss of performance when using simple margin with a linear kernel, a phenomena not observed for the other distributional models. This could indicate that the distributional representations produced by SGNS are locally noisy but globally coherent, and  representations produced by CBOW are locally coherent, but globally noisy.

It should also be noted that the training data used for the distributional models (BNC) is a comparably small, balanced, corpus. Results would be different for different sizes and kinds of corpora. 

\section{References}
\bibliographystyle{lrec}
\bibliography{main}

\end{document}